\documentclass[10pt,twocolumn,letterpaper]{article}

\usepackage[pagenumbers]{iccv} 

\definecolor{iccvblue}{rgb}{0.21,0.49,0.74}
\usepackage[pagebackref,breaklinks,colorlinks,allcolors=iccvblue]{hyperref}
\usepackage{multirow}
\usepackage[normalem]{ulem}
\useunder{\uline}{\ul}{}

\usepackage{colortbl}      

\def\paperID{7378} 
\def\confName{ICCV}
\def\confYear{2025}

\title{MUSE-VL: Modeling Unified VLM through Semantic Discrete Encoding}

\author{Rongchang Xie  ~~~~Chen Du ~~~~Ping Song ~~~~Chang Liu $^\dag$ \\
ByteDance\\
{\tt\small \{xierongchang, duchen.ai, songping.ldw\}@bytedance.com, wen8.zhou@gmail.com}
}

\begin{document}
\maketitle
\renewcommand{\thefootnote}{}
\footnotetext[0]{$^\dag$ Corresponding author}

\begin{abstract}
We introduce MUSE-VL, a Unified Vision-Language Model through Semantic discrete Encoding for multimodal understanding and generation. Recently, the research community has begun exploring unified models for visual generation and understanding. 
However, existing vision tokenizers (e.g., VQGAN) only consider low-level information, which makes it difficult to align with language tokens. This results in high training complexity and necessitates a large amount of training data to achieve optimal performance.
Additionally, their performance is still far from dedicated understanding models. This paper proposes \textbf{S}emantic \textbf{D}iscrete \textbf{E}ncoding (SDE), which effectively aligns the information of visual tokens and language tokens by adding semantic constraints to the visual tokenizer. This greatly reduces the amount of training data and improves the performance of the unified model. 
With the same LLM size, our method improved the understanding performance by 4.8\% compared to the previous SOTA Emu3 and surpassed the dedicated understanding model LLaVA-NeXT 34B by 3.7\%. For visual generation, our model achieves a FID score of 7.73 on MJHQ-30k, surpassing the existing unified models.

\end{abstract}
    
\section{Introduction}
\label{sec:intro}

Recently, there has been a growing interest in the domain of unified Multimodal Large Language Models (MLLMs). Researchers are dedicated to developing unified MLLMs by integrating visual understanding and generation tasks within autoregressive next-token prediction models. To realize a unified MLLM capable of next-token prediction for both visual and text tokens, one of the most critical challenges is \emph{how to convert visual input into discrete tokens, like text tokenizers do for text}.

\begin{figure}[ht]	
   \centering
    \begin{subfigure}{0.46\textwidth}
        \centering
        \includegraphics[width=\textwidth]{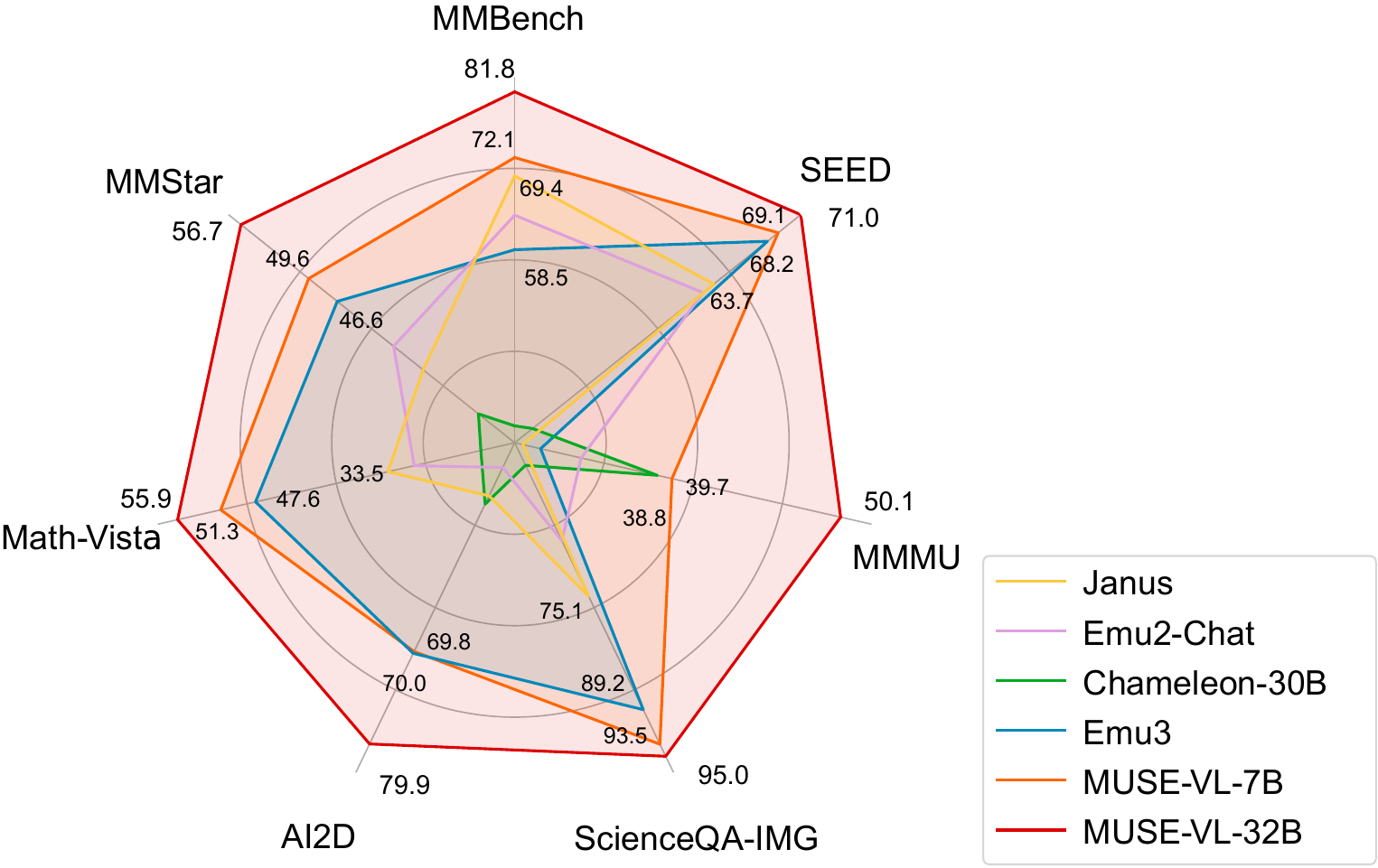}
        \label{fig:subfig1}
    \end{subfigure}
    \begin{subfigure}{0.46\textwidth}
        \centering
        \includegraphics[width=\textwidth]{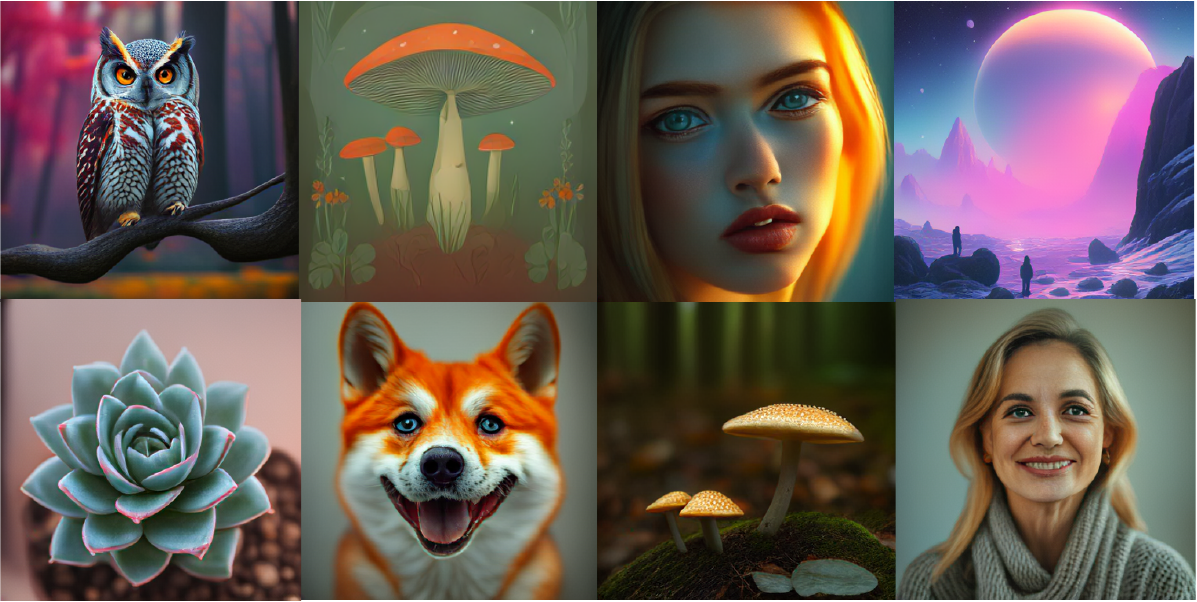}
        \label{fig:subfig2}
    \end{subfigure}
\caption{The evaluation of multimodal understanding and generation. \textbf{(Top)} Multimodal understanding results on various benchmarks. MUSE-VL surpasses the leading unified multimodal LLM, Emu3 \cite{wang2024emu3}. \textbf{(Bottom)} Images generated with MUSE-VL.}
\label{fig:res_show}
\end{figure}

\begin{figure*}[ht]
  \centering
   \includegraphics[width=\linewidth]{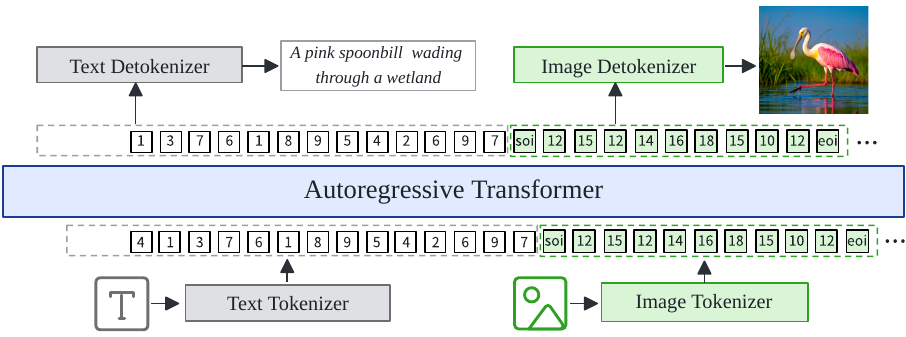}
   \caption{The overview of MUSE-VL. The image is converted into text-aligned visual tokens by the Semantic Discrete Encoder. The visual tokens and the corresponding text tokens, are fed into an autoregressive transformer. The training objective of the model is to predict the next token for both visual and text tokens. The predicted visual tokens can be decoded into an image by the Image Decoder. The \textit{soi} and \textit{eoi} tokens are used to mark the start and end of visual tokens, while the green tokens represent the visual tokens obtained through the visual tokenizer. 
   }
   \label{fig:main}
\end{figure*}

Some well done image discretization works have been processed in the past few years, such as VQ-VAE \cite{van2017neural}, VQ-GAN \cite{esser2021taming},  and MAGVIT \cite{yu2023language}. This enables the large language models (LLMs) to jointly learn visual code embeddings along with text tokens. However, previous unified works \cite{team2024chameleon, xie2024show} 
frequently demonstrate poor performance in multimodal understanding tasks, failing to match the performance of state-of-the-art multimodal understanding models \cite{liu2024llavanext}. This is mainly because these image quantization methods are trained solely with image reconstruction tasks and lack alignment with textual or semantic features. The discretization process is ineffective in capturing the high-dimensional semantic representation of images and inevitably leads to loss of semantic information. As a result, the visual tokens obtained by these methods are not suitable for visual understanding. As shown in Table \ref{table:ablation}, whether the visual tokens contain semantic information has a significant impact on visual understanding.

Recent research works attempt to address this issue. EMU3 \cite{wang2024emu3} reaches state-of-the-art results in both understanding and generation tasks by fine-tuning two separate models respectively, but they do not resolve this problem caused by unifying these two tasks within one model. Janus \cite{wu2024janus} uses separate encoders for understanding and generation, which increases the complexity of the model. Concurrent work TokenFlow \cite{qu2024tokenflow} designs a dual-codebook architecture to decouple the learning of semantic and pixel-level features. These methods all assume that images need to be represented using two different codebook spaces. In contrast, VILA-U \cite{wu2024vilau} combines contrastive loss and reconstruction loss to align the visual encoder with its textual input, but it struggles to converge in the additional semantic pretraining of image tokenizers, which is commonly attributed to loss conflicts. 

In this paper, we propose \textbf{S}emantic-aware \textbf{D}iscrete \textbf{E}ncoding (SDE) to avoid loss conflicts in VILA-U. Unlike VILA-U employs a text encoder to extract semantic features and then conducts contrastive learning, our SDE is learned from a pretrain CLIP-style teacher. It's worth noting that the features extracted by the image encoder of the pre-trained CLIP model \cite{radford2021learning, zhai2023sigmoid} already contain semantic information. Therefore, we use the image encoder of a pretrained CLIP-style model to extract semantic features. These semantic features, together with the image features from a visual encoder, are then used for quantization. As reconstructing from the quantized feature, we designed two decoder-like branches: an image decoder for image reconstruction and a semantic decoder to ensure that the discrete quantized features contain semantic information. Our approach considers semantic information during image discretization, meeting the requirements for both visual understanding and generation tasks. 

Building upon the SDE tokenizer, we introduce MUSE-VL, a state-of-the-art and easy-to-reproduce VLM. MUSE-VL is first pre-trained on image-text pairs to align language and visual tokens. The model is then fine-tuned with high-quality multimodal instruction-following data and visual generation data. The training objective is next-token prediction using the standard cross-entropy loss. The experiments demonstrate that MUSE-VL exhibits robust performance across complex multimodal tasks such as image reasoning, visual question answering, and image generation. As shown in Fig. \ref{fig:res_show}, MUSE-VL surpasses the current leading unified multimodal models \cite{team2024chameleon,wang2024emu3} on various benchmarks. 
The main contributions of this paper are as follows:

\begin{itemize}[left=1em]
\item  We develop a semantic-aware visual tokenizer SDE that can effectively integrate semantic features during the process of discretizing images into visual codes. This method allows seamless adaptation to any pre-trained LLM without modifications of model structure, facilitating the joint training of visual understanding and generation tasks.

\item  We propose MUSE-VL, a unified autoregressive transformer for multimodal understanding and generation. MUSE-VL models vision and language as unified discrete tokens and achieves state-of-the-art performance on various vision-language benchmarks.

\end{itemize}

\section{Related Work}
\label{sec:formatting}

\paragraph{Multimodal Understanding Models} A typical VLM for multimodal understanding can be abstracted into three modules, a pre-trained visual encoder \cite{radford2021learning, zhai2023sigmoid}, a pre-trained LLM \cite{touvron2023llama,yang2024qwen2,jiang2024mixtral,yang2023baichuan}, and a learnable connector \cite{liu2024improved, liu2023visual} between the visual encoder and LLM. Open-source VLMs have demonstrated strong multimodal understanding capabilities by aligning visual features of the pre-trained image encoder with the input embedding space of LLMs. VLMs can be roughly divided into two types based on the differences in how visual features are integrated into LLMs. One \cite{li2022blip,alayrac2022flamingo} injects visual information into LLMs using a cross-attention mechanism to fuse language and visual information. The other method \cite{liu2024visual,liu2024llavanext,young2024yi,bai2023qwen,wang2024qwen2,beyer2024paligemma,xu2024pllava,chen2024internvl,mckinzie2024mm1} directly appends the features extracted by the visual encoder with text embeddings to form the input sequence at the input layer of the LLMs.
Recently, the encoder-free models \cite{chen2024solo,diao2024eve,luo2025mono} aim to use the unified transformer architecture to address the challenges associated with encoder-based MLLMs, such as inflexible input and inefficient deployment.
SOLO \cite{chen2024solo} employs a single transformer architecture, which accepts raw image patches as inputs without a separate pre-trained vision encoder. 
EVE \cite{diao2024eve} proposes visual representation supervision and language conceptual alignment. 
Mono-InternVL \cite{luo2025mono} integrates a set of visual experts into a pre-trained LLM via a multimodal mixture-of-experts structure. 
In these VLMs, the visual features are continuous, which present challenges in the unified modeling of visual and language tokens.

\paragraph{Visual Tokenization for Generation}
Vector quantized (VQ)  visual tokenizers \cite{van2017neural,esser2021taming,yu2021vector,yu2023language,tian2025var} are proposed to convert image pixels into a sequence of discrete tokens and then reconstruct the input image from quantized features.  VQVAE \cite{van2017neural} first quantizes the image embeddings by performing a nearest neighbor look-up from the codebook, and then reconstructs the original image through a decoder. VQGAN \cite{esser2021taming} introduces a discriminator and perceptual loss to enhance the perceptual quality and details of the generated images. Recently, researchers have proposed residual quantization \cite{lee2022rq}, lookup-free quantization \cite{yu2023language} and multi-scale quantization \cite{tian2025var} to further improve the generation quality. However, these discrete VQ tokenizers are exclusively trained with the image reconstruction loss, without considering semantic features.

\paragraph{Unified Visual Language Models} Pioneering efforts have made significant strides by enabling multimodal understanding and generation within language models. In the realm of generating visual content using VLM, many works \cite{sun2023emu,ge2023making,zhan2024anygpt,jin2024unified,ge2024seedx,zhou2025transfusion, wang2024illume, huang2025illume+} have integrated VLMs with diffusion models \cite{rombach2022high} to achieve high-quality visual outputs. It is important to note that VLMs inherently lack the capability to directly produce visual content, and the quality of the generated images heavily relies on the performance of the diffusion models. For example, Emu \cite{sun2023emu} uses the output of the LLM as a condition for the pretrained diffusion model and then generates images with the diffusion model.  Transfusion \cite{zhou2025transfusion} combines the language modeling loss function with diffusion to train a single transformer.
ILLUME \cite{wang2024illume} proposes a self-enhancing multimodal alignment scheme to promote synergistic enhancement between understanding and generation capabilities.

Other works like Chameleon \cite{team2024chameleon}, Show-o \cite{xie2024show} and Emu3 \cite{wang2024emu3} have tried to directly adopt the VQ tokenizer to encode images for both multimodal understanding and generation.
However, since these visual tokenizers do not contain semantic information, aligning visual tokens with language tokens becomes difficult, and these models usually yield suboptimal performance in multimodal understanding tasks.

{VILA-U} \cite{wu2024vilau} combines contrastive and reconstruction loss to align visual and text tokens, but it has convergence problems, requiring a specific training recipe and large-scale image-text pairs from COYO-700M \cite{kakaobrain2022coyo-700m} dataset. 
SynerGen-VL \cite{li2025synergen} introduces the token folding mechanism and vision-expert-based progressive alignment pretraining strategy for building a unified MLLM. UniMoD \cite{mao2025unimod} proposes a task-aware token pruning method for the efficient training of MLLMs. 

 In this work, we explore a semantic-aware discrete encoding method for image reconstruction and generation. Our work reconstructs Siglip's visual features, which are well-aligned with text, making the training process simpler and demonstrating outstanding performance in both visual understanding and generation tasks.

\begin{figure*}[!htbp]
  \centering
   \includegraphics[width=1\linewidth]{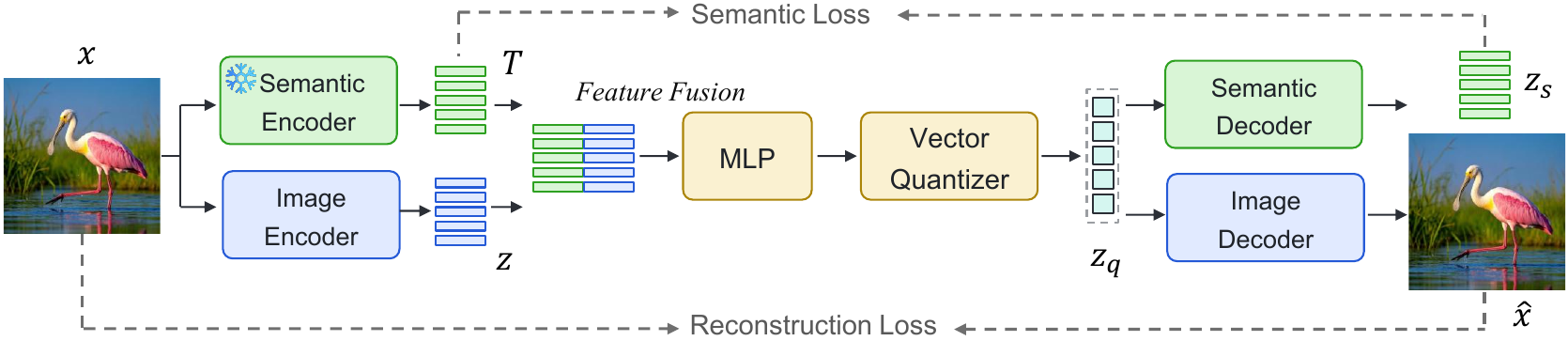}

   \caption{The overview of Semantic Discrete Encoding.
   The image is encoded and quantized into semantic discrete tokens, which are then separately reconstructed by the semantic decoder and the image decoder into semantic features and the original image.}
   \label{fig:onecol}
\end{figure*}
\section{Method}

The main objective of this work is to establish a simple and unified autoregressive transformer for both visual and language modalities. In this model, visual and language data can be input and output in the same discrete encoding format.  For the language modality, there are already well-developed text tokenizers and large language models (LLMs) \cite{yang2024qwen2,jiang2024mixtral,yang2023baichuan} that have been extensively trained on massive text data. However, how to construct an effective tokenizer in the visual modality remains to be explored.

Therefore, we propose semantic discrete encoding as a tokenizer for the visual modality to generate visual tokens that are well-aligned with language. Based on this, we propose MUSE-VL, a model capable of handling mixed visual and language tokens, supporting both visual understanding and generation tasks. This section first introduces the visual tokenizer proposed in our work, followed by the unified vision-language model built upon it.

\subsection{Visual Tokenizer}
\paragraph{Preliminary}
The VQGAN is a seminal work in the field of visual generation \cite{esser2021taming,van2017neural}. It learns a convolutional model consisting of an encoder $Enc$ and a decoder $Dec$, and it represents an image $x$ using codes $q$ from a learned, discrete codebook $\mathcal{Z} = \{{z}_k\}_{k=1}^K \subset \mathbb{R}^{d}$.

First, the image is encoded as $z = E(x) \in \mathbb{R}^{h \times w \times d}$ . Then, the index $q$ and quantized vector ${z}_q$ are obtained through element-wise quantization of each spatial feature $z$ onto its closest codebook entry. Finally, the decoder reconstructs the image from the quantized vector ${z}_q$.  The training objective consists of image reconstruction loss, vector quantizer (VQ) loss, discriminator
and perceptual loss. Further details can be found in the literature \cite{esser2021taming, sun2024autoregressive}.

\paragraph{Semantic Discrete Encoding (SDE)}
In this work, we propose the semantic discrete encoding method based on the Vector Quantizer. The architecture of the visual tokenizer is shown in Fig. \ref{fig:onecol}.
To guarantee that the discrete encoding produced by the tokenizer incorporates semantic information and is more closely aligned with language tokens, we introduce a semantic decoder and a semantic encoder. These components retrieve the semantic information of the image from the discrete codes.

Specifically, for an image $ x \in \mathbb{R}^{H \times W \times 3} $, where $ H $ and $ W $ represent height and width dimensions, respectively, the image encoder produces the feature $ z = {Enc}(x) $. Here, $ z \in \mathbb{R}^{h \times w \times d} $, where  $h$ and $w$ are the sizes after downsampling, and $d$ is the codebook vector dimension. 
Subsequently, the feature $ z $ is transformed into quantization code $ q \in \mathbb{Z}^{h \times w}$ and quantized embedding $ z_q \in \mathbb{R}^{h \times w \times d} $ through the quantization operation. To ensure that the quantized embedding carries meaningful semantics, inspired by BEITv2 \cite{peng2022beit}, a transformer is used as the semantic decoder $Dec_s$ to maximize the cosine similarity between the decoded feature $ z_s $ and a pre-trained semantic feature $ T $:
\[
L_{\text{sem}} = 1 - \cos(z_s, T) = 1 - \cos({Dec}_s(z_q), T)
\]

In our study, we adopt the SigLIP model \cite{zhai2023sigmoid} as the semantic encoder to produce semantic feature $ T $, which has been trained with an extensive dataset of image-text pairs and has been effectively aligned with language. The parameters of the semantic encoder are frozen in the training.
To further enhance the semantics of discrete coding, we fuse the semantic feature $T$ with the image feature $z$ in the encoding process, and then quantize the merged feature:
\[
z^q = \text{Quant}(T + z)
\]

Additionally, to preserve the image generation capability of the tokenizer, a separate image decoder is used to generate the reconstructed image $\hat{x}$. Consequently, the final loss function is a combination of semantic loss \(L_{\text{sem}}\), image reconstruction loss, and VQ loss in VQGAN: 

\[
L_{\text{total}} = L_{\text{sem}} + L_{\text{img}} + L_{\text{vq}}
\]
where,
\[
L_{\text{img}} = \ell_2(x, \hat{x}) + L_P(x, \hat{x}) + \lambda_G L_G(\hat{x})
\]
\[
L_{\text{vq}} = \| \text{sg}[z] - z_q \|_2^2 + \beta \| z - \text{sg}[z_q] \|_2^2
\]

Here, $\ell_2$ is the L2 reconstruction loss, $L_P(\cdot)$ refers to the perceptual loss measured by LPIPS \cite{zhang2018unreasonable}, and $L_G(\cdot)$ is an adversarial loss \cite{isola2017image}. The second term of $L_{\text{vq}}$ is the commitment loss \cite{van2017neural}, and $\beta$ is its weight. We used a convolutional network as the image decoder, following VQGAN \cite{esser2021taming,sun2024autoregressive}.

\subsection{Unified Vision-Language Modeling}
Based on semantic discrete encoding, we propose a unified vision-language modeling named MUSE-VL. The structure of MUSE-VL is shown in the Fig. \ref{fig:main}. The image is pre-processed into visual tokens $\{q_1, q_2, ..., q_i\}$ of length $(h \times w)$ through SDE tokenizer, while the textual data is converted through the text tokenizer. To achieve joint modeling of language and vision, it is sufficient to simply extend the embedding layer of existing LLMs to incorporate newly added visual token IDs. This modification enables seamless integration of multimodal inputs within the model's architecture. To distinguish visual tokens, two special tokens \textless $soi$\textgreater and \textless $eoi$\textgreater are added to mark the start and end of visual tokens respectively. The training objective of the model remains a simple autoregressive task, without any modifications to the LLM's architecture or training loss.

In this work, we adopt Yi-1.5 \cite{young2024yi} and Qwen-2.5 \cite{qwen2.5, qwen2}, which perform well on language tasks, as the base LLM. It is crucial to emphasize that the inherent alignment of SDE tokenzier with language and the unified autoregressive architecture enables our model to integrate effortlessly with the most LLMs. This integration is achieved using only minimal image-text data and does not require any architecture modifications. In contrast, previous approaches, such as Chameleon and Emu3 \cite{team2024chameleon, wang2024emu3}, necessitated alterations to the model architecture and required extensive image and language data to train the LLM from scratch.

\paragraph{Pretraining}
In the pre-training stage, we used images with paired text descriptions for training. At this stage, we calculated the loss for all tokens to optimize the model parameters using a next-token prediction objective. The primary objective at this stage was to effectively learn a robust embedding of visual tokens, align visual tokens with text tokens, and build the model's capability to accurately predict image tokens.

\paragraph{Instruction Tuning}
For image understanding tasks, our work uses visual instruction tuning data and image caption data. These data are organized in the following format, where the visual tokens appear in the prompt, and the target part is the response text. Only tokens in the target part participate in the loss calculation:

\textit{
Prompt: \{text\} \textless soi\textgreater  \{vision  tokens\} \textless eoi\textgreater 
}

\textit{
Target:  \{response\}
}

For the image generation task, the order of images and texts in the image caption data is reversed here, enabling the model to generate visual tokens based on the descriptions.

\textit{
Prompt: \{system text\} \{image caption\} }

\textit{
Target:   \textless soi \textgreater  \{vision tokens\} \textless eoi \textgreater}

The system text is randomly sampled from a set of image generation instructions, such as ``Please generate an image.'', ``Show me a photo.'', etc. At the inference stage, the user provides a prompt for generating an image, and the model will predict the corresponding image tokens. Then, the predicted visual tokens are converted to the image by the image decoder.

\section{Experiments}

\subsection{Implementation Details}

\begin{table}[!ht]
\centering
\caption{Comparison of different visual tokenizers on multimodal understanding benchmarks. All models used the same base LLM and training dataset.}
\label{tab:tokenizer-yi_compare}
\setlength{\tabcolsep}{4pt}
\begin{tabular}{@{}lcccccc@{}}
\toprule
\textbf{Method}  & \textbf{MMBench}  & \textbf{SEED}  & \textbf{MMStar} & \textbf{AVG} \\
\midrule
VQGAN \cite{esser2021taming}  & 32.0  & 42.7  & 29.1 & 34.6 \\
SEED \cite{ge2023planting}  & 63.1 & 57.8  & 39.1 & 53.3 \\
LaVIT \cite{jin2024unified}  & 63.3  & 59.5  & 40.3 & 54.4\\
\rowcolor{gray!20} 
\textbf{SDE (ours)} & \textbf{70.6}  & \textbf{68.1} & \textbf{43.8} &\textbf{60.8} \\ \bottomrule

\end{tabular}
\end{table}

\begin{table*}[ht]
\centering
\small
\caption{Evaluation on multimodal understanding benchmarks. Compared with previous methods, our MUSE-VL achieved the best performance on various benchmarks. The best results for unified models with fewer than 10B parameters are in bold, while the best of all unified models are underlined.
}
\label{tab:muse-vl_compare_vlms}
\setlength{\tabcolsep}{2.0pt}
\begin{tabular}{lllcccccccccc}
\toprule

\textbf{Method} & \textbf{LLM}         & \textbf{Visual Token} & \textbf{Res.} & \textbf{MMBench} & \textbf{MMStar} & \textbf{SEED} & \textbf{MMMU} & \textbf{SQA-I} & \textbf{AI2D} & \textbf{MathVista} & \textbf{AVG}  \\
\midrule
\multicolumn{12}{l}{\textit{Understanding Only}}  \\
\midrule
InstructBLIP \cite{InstructBLIP}    & Vicuna-7B            & Continuous            & 224                 & 36.0             & 32.7            & 58.8          & 30.6          & 60.5                    & 33.8          & 24.4                & 39.5          \\
LLaVA-1.5 \cite{liu2024improved}      & Vicuna-1.5-7B        & Continuous            & 336                 & 64.3             & 33.1            & 66.1          & 35.7          & 66.8                    & 55.5          & 27.4                & 49.8          \\
LLaVA-NeXT \cite{liu2024llavanext}     & Vicuna-1.5-7B        & Continuous            & 672                 & 67.4             & 37.6            & {70.2} & 35.8          & 70.1                    & 66.6          & 34.6                & 54.6          \\
LLaVA-NeXT \cite{liu2024llavanext}     & Yi-34B & Continuous            & 672                 & 79.3             & 51.6            & { 75.9}    & { 51.1}    & 81.8                    & 78.9          & 46.5                & 66.4          \\
ShareGPT4V \cite{chen2023sharegpt4v}     & Vicuna-1.5-7B        & Continuous            & 336                 & 68.8             & 35.7            & 69.7          & 37.2          & 68.4                    & 58.0          & 26.5                & 52.0          \\
VILA  \cite{lin2024vila}          & LLaMA-2-7B           & Continuous            & 336                 & 68.9             & -               & 61.1          & -             & 68.2                    & -             & -                   & -             \\
EVE \cite{diao2024eve}           & Vicuna-7B      & Pixel        & -   & 52.3 & -    & 56.8 & -    & 64.9 & -    & -    & - \\
SOLO  \cite{chen2024solo}        & Mistral-7B     & Pixel        & 1024   & -    & 35.5 & 64.4 & -    & 73.3 & 61.4 & 34.4 & -    \\
Mono-internvl \cite{luo2025mono} & InternLM2-1.8B & Pixel        & -   & 65.5 & -    & 67.4 & 33.7 & 93.6 & 68.6 & 45.7 & - \\
Qwen2.5 VL \cite{bai2025qwen25vl}  & Qwen2.5-7B  & Continuous & - & 83.5 & 63.9 & 77.0 & 58.6 & 88.9 & 83.9 & 68.2 & 74.9 \\
Qwen2.5 VL \cite{bai2025qwen25vl} & Qwen2.5-72B & Continuous & - & 88.6 & 70.8 & 79.5 & 70.2 & 91.3 & 88.7 & 74.8 & 80.6 \\
\midrule
\multicolumn{12}{l}{\textit{Understanding and Generation}}  \\
\midrule
DreamLLM \cite{dong2024dreamllm}        & Vicuna-7B            & Continuous            & 224                 & 58.2             & -               & -             & -             & -                       & -             & -                   & -             \\
Unified-IO2 \cite{lu2024unified}     & 7B from scratch    & Continuous            & 384                 & 71.5             &  -               & 61.8          & -             & 86.2                    & -             & -                   & -             \\
Emu2-Chat \cite{sun2024generative}      & LLaMA-33B            & Continuous            & 448                 & 63.6             & 40.7            & 62.8          & 34.1          & 68.2                    & 49.7          & 30.7                & 50.0          \\
Video-LaVIT \cite{jin2024video}    & Llama 2 7B           & Continuous            & 224                 & 67.3             & -               & 64.0          & –             & 70.0                    & -             & -                   & -             \\
Janus \cite{wu2024janus}          & DeepSeek-1.3B    & Continuous            & 384                 & 69.4             & 37.6            & 63.7          & 30.5          & 75.1                    & 52.8          & 33.5                & 51.8          \\
\midrule
Chameleon \cite{team2024chameleon}   & 7B from scratch      & Discrete              & 512                 & 31.1             & 31.1            & 30.6          & 25.4          & 46.8                    & 46.0          & 22.3                & 33.3          \\
Chameleon \cite{team2024chameleon}  & 34B from scratch     & Discrete              & 512                 & 32.5             & 31.8            & 48.5          & 38.8          & 58.8                    & 53.7          & 23.6                & 41.1          \\
SEED-LLaMA \cite{li2023seed}      & Vicuna-7B            & Discrete              & 224                 & 28.7             & 33.1            & 51.5          & -             & -                       & -             & -                   & -             \\
Show-o  \cite{xie2024show}        & Phi-1.5-1.3B         & Discrete              & 256                 & -                & -               & -             & 25.1          & -                       & -             & -                   & -             \\
VILA-U \cite{wu2024vilau}         & LLaMA-2-7B           & Discrete              & 384                 & -                & -               & 59.0          & -             & -                       & -             & -                   & -             \\
SynerGen-VL \cite{li2025synergen}   & InternLM2-1.8B & Discrete & -   & 53.7 & -    & 62.0 & 34.2 & 92.6 & 60.8 & 42.7 & - \\
UniMoD \cite{mao2025unimod}      & 8B             & Discrete & 512 & -    & -    & -    & 25.3 & -    & -    & -    & -   \\
Emu3  \cite{wang2024emu3}          & 8B from scratch      & Discrete              & 512                 & 58.5             & 46.6               & 68.2          & 31.6          & 89.2                    & \textbf{70.0} & 47.6                   & 58.8              \\
\rowcolor{gray!20} 
MUSE-VL (ours)  & Qwen2.5-7B           & Discrete              & 256                 & \textbf{72.1}  &  \textbf{49.6} & \textbf{69.1} & \textbf{39.7} & \textbf{93.5} & 69.8       & \textbf{51.3} & \textbf{63.6} \\
\rowcolor{gray!20} 
MUSE-VL (ours)  & Qwen2.5-32B          & Discrete              & 384                 & {\ul 81.8}       & {\ul 56.7}      & {\ul 71.0}          & {\ul 50.1  }        & {\ul 95.0}              & {\ul 79.9}    & {\ul 55.9}          & {\ul 70.1}   \\
\bottomrule

\end{tabular}
\end{table*}

\paragraph{Visual Tokenizer}
For the pre-trained semantic encoder, this work uses two different resolution encoders: SigLIP-SO400m-patch14-384 and SigLIP-Large-patch16-256 \cite{zhai2023sigmoid}. The semantic decoder is a vision transformer (same as BEITv2 \cite{peng2022beit}). The input images are resized to 384 $\times$ 384 and 256 $\times$ 256 respectively, and after quantization, they are converted into discrete codes of 16 $\times$ 16 and 27 $\times$ 27. 

We use the pretrained SigLIP parameters as the initialization of the image encoder in SDE, while the image decoder remains the same ConvNet architecture as the VQGAN decoder \cite{esser2021taming, sun2024autoregressive}. The codebook size of the tokenizer is 32,768, with a vector dimension of 8, and the semantic loss weight is set to 1. The training dataset for the tokenizer includes 10 million images from ImageNet-1K \cite{deng2009imagenet} and CC12M \cite{changpinyo2021conceptual} . Other hyperparameters during training follow the default settings in LLamaGEN \cite{sun2024autoregressive}.

\paragraph{Vision Language Model}
The method proposed in this paper can be easily adapted to most pre-trained LLMs. In our experiments, we used Qwen-2.5-7B, Qwen-2.5-32B \cite{qwen2.5}, Yi-1.5-9B, and Yi-1.5-34B \cite{young2024yi} as the base language models. The embedding layer of the LLM is expanded by 32,768 to accommodate visual tokens. 
The learning rate during training is set to 1e-4, using a cosine schedule with warmup, and AdamW ($\beta1$ = 0.9, $\beta2$ = 0.95) as the optimizer.
We use the image caption dataset LLaVA-ReCap-CC12M \cite{li2024onevision} for the pre-training stage.
For visual understanding tasks, we used examples from Cambrian7M \cite{tong2024cambrian} and LLaVA-OneVision-Data \cite{li2024onevision}. For visual generation tasks, we used datasets \cite{changpinyo2021conceptual} and 10M high-quality images.
The images for visual generation are resized while maintaining the original aspect ratio.

\paragraph{Evaluation Setup}
To evaluate the multimodal understanding capability, we use benchmarks such as MMBench \cite{mmbench}, SeedBench-img \cite{li2023seed}, AI2D \cite{kembhavi2016diagram}, MMStar \cite{chen2024mmstar}, MathVista \cite{lu2024mathvista}, SciQA-Img \cite{lu2022sqa}, TextVQA \cite{singh2019towards} and MMMU \cite{yue2024mmmu}. 
We run the evaluation based on VLMEvalKit \cite{duan2024vlmevalkit}.
To evaluate the visual generation capability, we use the MJHQ-30K \cite{li2024playground} and GenEval \cite{2023geneval} benchmarks. In MJHQ-30K, the quality of the generated images is measured by calculating the Fréchet Inception Distance (FID) \cite{heusel2017gans} between 30K generated samples and high-quality samples. The GenEval \cite{2023geneval} benchmark is used for evaluating the model's text-to-image alignment.

\subsection{Evaluation of Visual Tokenizer}
 \paragraph{Comparison with other Visual Tokenizers}
We validate the impact of different visual tokenizers on the performance of VLMs. 
Table \ref{tab:tokenizer-yi_compare} summarizes the comparison between the proposed tokenizer and other visual tokenizers across various multimodal understanding benchmarks. It should be noted that all results were obtained using the same LLM (Yi-1.5-9B) and the same subset of the training set. 

We use the pre-trained VQGAN model from LLamaGEN \cite{sun2024autoregressive} as the baseline method, which has excellent performance in image reconstruction and generation. 
Table \ref{tab:tokenizer-yi_compare} shows that VQGAN performs poorly in multimodal understanding tasks due to difficulties in aligning with text, and we also found that this model often misidentifies the object of the image. 
By considering semantic information in the visual discretization process, the proposed SDE tokenizer  extracts visual tokens that are more aligned with text, thus exhibiting strong multimodal understanding capabilities.

We also compare our method with other discrete visual tokenizers. The SEED tokenizer proposed by SEEDLLaMA \cite{ge2023making} optimizes image tokens for both discriminativeness and reconstruction at the training stage. LaVIT \cite{jin2024unified} introduces a dynamic visual tokenizer. We replace the LLM in LaVIT with Yi-1.5-9B for a fair comparison. 

As shown in Table \ref{tab:tokenizer-yi_compare}, compared to the recent works SEED \cite{ge2023making} and LaVIT's tokenizer \cite{jin2024unified}, the proposed tokenizer exceeds by a large margin (+7.5\% and +6.4\%) in accuracy. The results indicate that the proposed method is more effective than other tokenizers in enhancing the multimodal understanding capabilities of VLM.

\begin{table}[!t]
\centering
\caption{Evaluation of visual tokenizer on image reconstruction. The evaluations are on ImageNet 50k validation set under the image resolution of $256\times 256$. }
\label{tab:tokenizer_compare}
\setlength{\tabcolsep}{2pt}
\begin{tabular}{lccccc}
\toprule
\textbf{Method} & \textbf{Code Size} & \textbf{Dim} & \textbf{rFID↓} & \textbf{PSNR↑} & \textbf{SSIM↑} \\ 
\midrule

VILA-U \cite{lee2022rq} & 1024 & - & 1.80 & - & - \\
\midrule
VQGAN \cite{esser2021taming} & 256 & 256 & 4.99 & 20.00 & 0.629 \\
RQ-VAE \cite{lee2022rq} & 256 & 256 & 3.20 & - & - \\
MaskGIT \cite{chang2022maskgit} & 256 & 256 & 2.28 & - & - \\
LLamaGen \cite{sun2024autoregressive} & 256 & 8 & 2.19 & 20.79 & 0.675 \\
\rowcolor{gray!20} 
SDE (ours) & 256 & 8 & 2.26 & 20.14 & 0.646 \\ \bottomrule

\end{tabular}
\end{table}

 \paragraph{Image Reconstruction}

 Table \ref{tab:tokenizer_compare} presents the quantitative results of the tokenizer on image reconstruction. We use r-FID (reconstruction-Fréchet Inception Distance), PSNR (Peak Signal to Noise Ratio), and SSIM (Structural Similarity) as metrics for assessing image reconstruction on the ImageNet 50k validation set.

 As summarized in table \ref{tab:tokenizer_compare}, the SDE tokenizer matches the state-of-the-art method LLamaGEN \cite{sun2024autoregressive} and surpasses VQGAN \cite{esser2021taming}, RQ-VAE \cite{lee2022rq}, and MaskGIT \cite{chang2022maskgit}. It is worth noting that our method needs to consider both semantic and image reconstruction simultaneously. In contrast, most previous methods focused solely on image reconstruction. 
 We achieved a similar rFID compared to VILA-U \cite{wu2024vilau}, even though its code size is four times ours.

\begin{figure*}[htbp]
  \centering
   \includegraphics[width=0.9\linewidth]{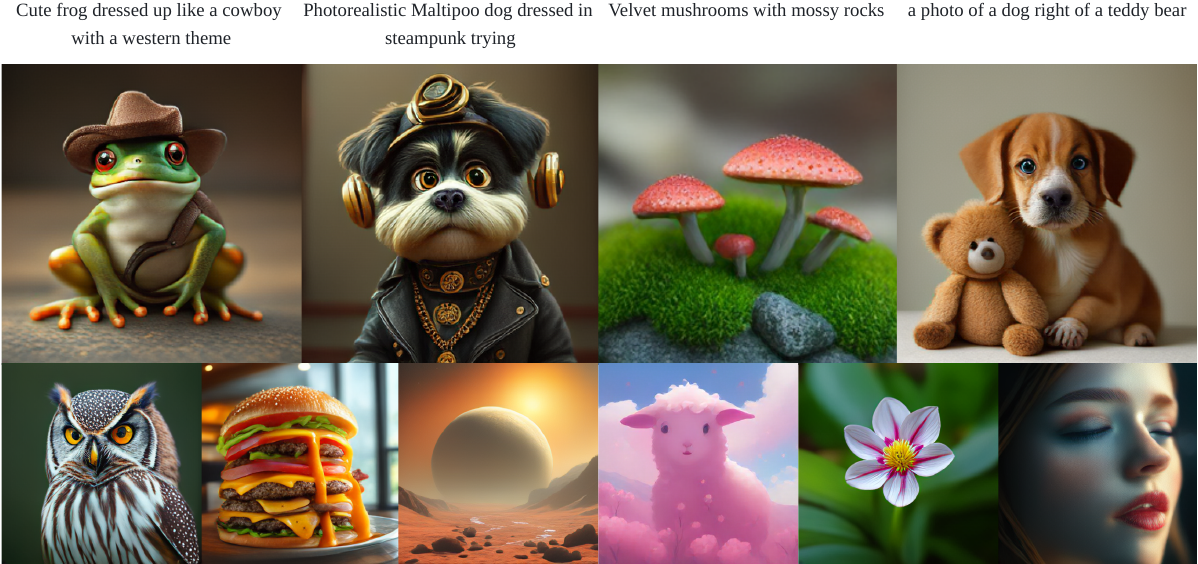}
   \caption{The generated images from MUSE-VL 7B. }
   \label{fig:t2i}
\end{figure*}

\subsection{Evaluation of Vision Language Model}
 \paragraph{Multimodal Understanding}
Table \ref{tab:muse-vl_compare_vlms} shows the comparison between MUSE-VL and other leading VLMs on various multimodal understanding benchmarks. We include understanding models and unified models for understanding and generation. We categorize the methods into three types: discrete VLMs, continuous VLMs and encoder-free VLMs, depending on whether the input visual features are continuous embeddings, discrete tokens, or raw pixel values. Table \ref{tab:muse-vl_compare_vlms} shows that discrete-visual-token models often perform worse than continuous-visual-embedding models, mainly due to challenges in aligning visual tokens with text tokens. Thanks to the proposed semantic discrete encoding tokenizer (SDE), the proposed MUSE-VL outperforms other discrete-visual-token VLMs and achieves better or comparable performance compared with continuous-visual-embedding VLMs. MUSE-VL with 7B parameters reaches 72.1\% on the MMBench, +13.6\% higher than the previous SOTA discrete method Emu3 \cite{wang2024emu3} and other models with the same parameter size.

MUSE-VL with 32B parameters achieves state-of-the-art results in unified models, which exhibit remarkable scalability of the proposed method.

\begin{table}[]
\centering
\small
\caption{Comparison of the number of image-text pairs in the training set. 
}
\label{tab:number}
\setlength{\tabcolsep}{4.5pt}
\begin{tabular}{l|lll}
\toprule
\textbf{Method} & Chameleon \cite{team2024chameleon} & SEED-LLaMA \cite{li2023seed}& Janus \cite{wu2024janus}    \\
\midrule
\textbf{Number}    & 1.4B      & 600M       & 65M       \\
\toprule
\textbf{Method} & Show-o \cite{xie2024show}   & VILA-U \cite{wu2024vilau}    & Ours \\
\midrule
\textbf{Number}    & 35M       & 720M       & 24M      \\
\bottomrule
\end{tabular}
\end{table}

The table \ref{tab:number} shows the number of image-text pairs used in unified multimodal models. Previous unified models such as Chameleon, SEED-LLaMA and VILA-U, typically rely on extensive image-text pairs to align visual and language tokens. By reconstructing the semantic features, the alignment and training process of VLM becomes more efficient, surpassing other models with only 24M data.

\begin{table}[]
\centering
\small
\caption{Quantitative results on text-to-image benchmarks. $\dagger$ result is with rewriting.}
\label{tab:t2ibench}
\setlength{\tabcolsep}{2pt}
\begin{tabular}{llccc}
\toprule
Type                           & Method          & Res. & MJHQ-30K ↓ & GenEval   \\
\midrule
\multirow{6}{*}{Gen. Only}     & SDv1.5 \cite{2022sd}          & 512  & -             & 0.43      \\
                                & PixArt \cite{chen2023pixartalpha}         & 512  & 6.14          & 0.48      \\
                               & SD-XL \cite{podell2024sdxl}          & 1024 & 9.55          & 0.55      \\
                               
                               & Play v2.5 \cite{li2024playground} & 1024 & 4.48          & -         \\
                               & DALL-E3  \cite{2023dalle3}      & 1024 & -             & 0.67$^\dagger$     \\
                               & LlamaGen  \cite{sun2024autoregressive}      & 512  & -             & 0.32      \\
\hline
\multirow{6}{*}{Und. and Gen.} & SEED-X  \cite{ge2024seedx}        & 1024 & -             & 0.49      \\
                                & Chameleon  \cite{team2024chameleon}  & 512  & -         & 0.39      \\
                                & LWM  \cite{liu2024world}           & 256  & 17.77         & 0.47      \\
                               
                               & Show-o  \cite{xie2024show}       & 256  & 15.18         &  \underline{0.53}      \\
                               & Janus  \cite{wu2024janus}         & 384  & \underline{10.10}         & \textbf{0.61}      \\
                               & VILA-U  \cite{wu2024vilau}        & 256  & 12.81         & -         \\
                               \rowcolor{gray!20} 
                               & Ours (7B)      & 256  & \textbf{7.73}          & \underline{0.53} / 0.57$^\dagger$ \\
\bottomrule
\end{tabular}
\end{table}

 \paragraph{Visual Generation}

Table \ref{tab:t2ibench} shows the quantitative results of the text-to-image in GenEVAL \cite{2023geneval} and MJHQ-30K \cite{li2024playground}. We compare MUSE-VL with other state-of-the-art generation-only models and unified models.
As shown in Table \ref{tab:t2ibench}, MUSE-VL achieves a 7.73 FID score on the MJHQ30K benchmark, which outperforms previous SOTA unified models and SD-XL. This demonstrates that our model can generate images with high aesthetics and quality. The Geneval results show that our model achieves better or comparable performance compared to other unified models, indicating that the generated images align well with the text prompts.
Figure \ref{fig:t2i} presents examples of visual generation.

\subsection{Ablation Studies}
 \paragraph{Effect of Semantic Branch and Image Branch}

Table \ref{table:ablation} presents the ablation study of the SDE tokenizer, validating the impact of the semantic and image branches on image reconstruction and understanding capabilities. 
We use the rFID on the ImageNet validation set to evaluate reconstruction capabilities, and MMB, SEED and MMStar to evaluate understanding capabilities.
The baseline tokenizer consists of an image encoder and an image decoder, with the training task being image reconstruction. It shows that the baseline performs poorly on the multimodal understanding task, which confirms the limitations of VQ tokenizers due to the pixel-level reconstruction focusing on low-level features. 

When only the semantic reconstruction task is performed (Row 2), using a semantic encoder and semantic decoder, there is a significant improvement in the understanding capability, demonstrating the importance of semantic representation for understanding tasks. However, the tokenizer lacks image reconstruction ability and cannot decode discrete tokens into images.
The SDE tokenizer (Row 3), by simultaneously reconstructing image and semantic features, integrates high-level and low-level information during the image discretization process. Compared to the baseline, it significantly improves visual understanding performance by 20.5\% and reduces the image reconstruction rFID.

\begin{table}[t]
    \setlength{\tabcolsep}{2.5pt}
    \centering
	\caption{Ablation study of the semantic and image branches in SDE tokenizer. The rFID represents reconstruction capability, while the rest represent multimodal understanding capability.}
    \label{table:ablation}
\begin{tabular}{cc|c|cccc}
\toprule
\textbf{Image} & \textbf{Semantic} & \textbf{rFID} & \textbf{MMB}    & \textbf{SEED} & \textbf{MMStar} & \textbf{AVG} \\

\midrule
\checkmark              &              & 2.63   & 42.8  & 48.5        & 38.1     & 43.1      \\
            &  \checkmark               & -      & \textbf{72.5}  & 67.5        & 48.1     & 62.7      \\
  \checkmark            &  \checkmark               & \textbf{2.26}   & 72.1  & \textbf{69.1}        & \textbf{49.6}     & \textbf{63.6}  \\
\bottomrule
\end{tabular}
\end{table}

\begin{table}[t]
\centering
\caption{Ablation of MUSE-VL on LLM and image resolution.}
\label{tab:muse-vl_compare}
\setlength{\tabcolsep}{3.5pt}
\begin{tabular}{llccccc}
\toprule
\textbf{LLM} & \textbf{Res} & \textbf{MMB} & \textbf{SEED} & \textbf{MMStar} & \textbf{AVG} \\

\midrule
Yi-1.5-9B    & 256        & 70.6         & 66.1          & 43.8            & 60.2         \\
Yi-1.5-9B    & 384        & 73.2         & 69.2          & 47.4            & 63.3         \\
Yi-1.5-34B   & 256        & 73.5         & 67.3          & 48.9            & 63.2         \\
\midrule
Qwen-2.5-7B  & 256        & 71.0         & 65.8          & 44.2            & 60.3         \\
Qwen-2.5-32B & 256        & 75.1         & 65.7          & 50.3            & 63.7        \\

\bottomrule
\end{tabular}
\end{table}

 \paragraph{Ablation on LLM and Resolution}
In this section, we use two series of LLMs (Yi and Qwen) as the base model of MUSE-VL and investigate the impact of LLM and image resolution on understanding performance. All models were trained using the same subset of the training set. The results are shown in Table \ref{tab:muse-vl_compare}.
Firstly, we found that for both the Yi and Qwen series, larger models consistently yield better results, confirming that the VLM architecture adheres to the scale-up theory. Secondly, for the Yi series, a larger input size (384 vs 256) also leads to improved performance.
The results show that MUSE-VL exhibits outstanding adaptability and scalability, with a better base LLM and larger model size consistently leading to superior performance.

\section{Conclusion}
This study presents SDE, a semantic-aware discrete encoding method devised to unify the input formats of images and texts within VLMs. The experimental results show that the SDE tokenizer is effective for VLMs handling both visual comprehension and generation tasks. Building upon the proposed semantic-aware visual tokenizer, we propose MUSE-VL, a unified vision-language model. This innovative model integrates both image and language understanding and generation tasks within a unified autoregressive next-token prediction framework. Our method is more efficient than existing unified VLMs and it demonstrates that the discrete autoregressive method can achieve comparable or even better performance than other advanced VLMs. 

\paragraph{Acknowledgment}
We thank Yuzhong Wang, Xibin Wu, Cheng Chen, and Tuoyu Zhang for their contributions to the infrastructure and the data processing pipeline.

{
    \small
    \bibliographystyle{ieeenat_fullname}
    \bibliography{main}
}

\clearpage

\maketitlesupplementary

\section*{Appendix}
\setcounter{section}{0}
\renewcommand\thesection{\Alph{section}}

\begin{table*}[!htbp]
\begin{tabular}{llccccccc}
\toprule
Type                           & Method           & Overall & Single Obj. & Two Obj. & Counting & Colors & Position & Color Attri. \\
\midrule
\multirow{7}{*}{Gen. Only}     & DALL-E 2 \cite{ramesh2022hierarchical}        & 0.52    & 0.94        & 0.66     & 0.49     & 0.77   & 0.1      & 0.19         \\
                               & SDv1.5 \cite{2022sd}         & 0.43    & 0.97        & 0.38     & 0.35     & 0.76   & 0.04     & 0.06         \\
                               & SDv2.1  \cite{2022sd}         & 0.50     & 0.98        & 0.51     & 0.44     & 0.85   & 0.07     & 0.17         \\
                               & SDXL  \cite{podell2024sdxl}           & 0.55    & 0.98        & 0.74     & 0.39     & 0.85   & 0.15     & 0.23         \\
                               & PixArt-alpha \cite{chen2023pixartalpha}    & 0.48    & 0.98        & 0.5      & 0.44     & 0.8    & 0.08     & 0.07         \\
                               & DALL-E 3  \cite{2023dalle3}       & 0.67 $\dagger$   & 0.96        & 0.87     & 0.47     & 0.83   & 0.43     & 0.45         \\
                               & LlamaGen \cite{sun2024autoregressive}       & 0.32    & 0.71        & 0.34     & 0.21     & 0.58   & 0.07     & 0.04         \\
\midrule
\multirow{5}{*}{Und. and Gen.} & Chameleon \cite{team2024chameleon} & 0.39    & -        & -     & -     & -    & -     & -         \\
                                & LWM \cite{liu2024world}             & 0.47    & 0.93        & 0.41     & 0.46     & 0.79   & 0.09     & 0.15         \\
                              
                               & SEED-X \cite{ge2024seedx}          & 0.49    & 0.97        & 0.58     & 0.26     & 0.8    & 0.19     & 0.14         \\
                               & Show-o \cite{xie2024show}          & 0.53    & 0.95        & 0.52     & 0.49     & 0.82   & 0.11     & 0.28         \\
                               \rowcolor{gray!20} 
                               & Ours (7B) & 0.53 & 0.99 & 0.65 & 0.44 & 0.73 & 0.18 & 0.17 \\
                               \rowcolor{gray!20} 
                               & Ours (7B) & 0.57 $\dagger$ & 0.98 & 0.64 & 0.52 & 0.72 & 0.25 & 0.31 \\
\bottomrule
\end{tabular}
\caption{Evaluation of text-to-image generation on the GenEval \cite{2023geneval}. $\dagger$ result is with rewriting.}
\label{fig:geneval}
\end{table*}

\section{Additional Results}
\begin{figure}[!htbp]
  \centering
   \includegraphics[width=\linewidth]{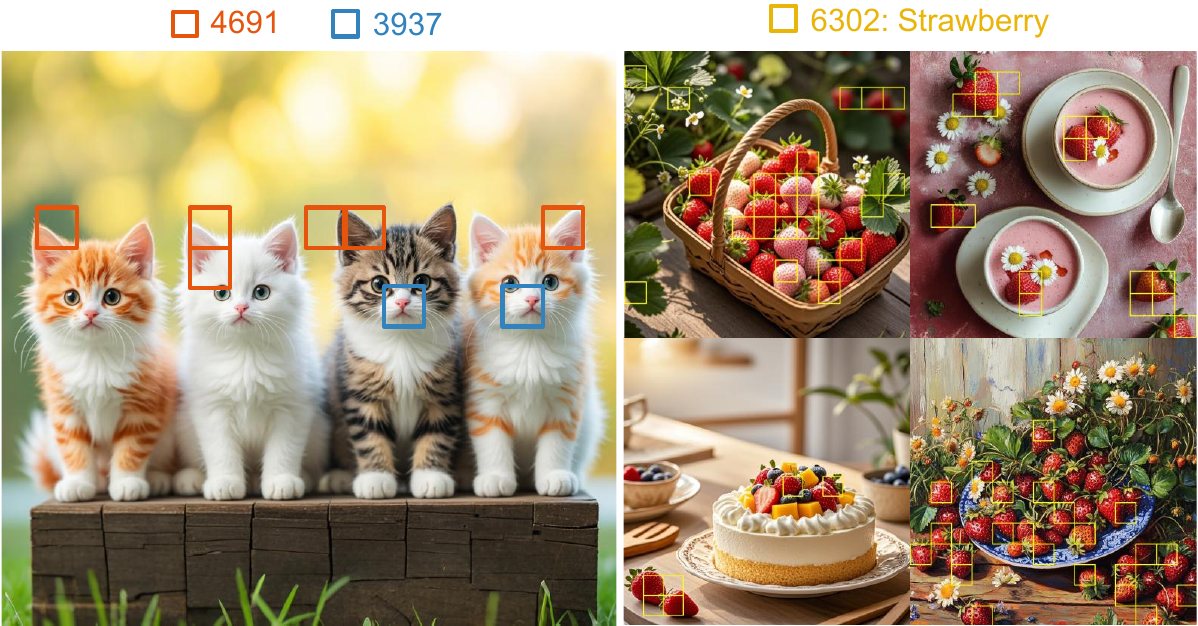}
   \caption{Visualization of semantic discrete codes. Rectangular boxes of the same color indicate that the corresponding semantic ID of these patches is the same. It can be observed that semantic ID can represent a semantic concept.} 
   \label{fig:id_vis}
\end{figure}

\paragraph{Visualization of Semantic Code}
Figure \ref{fig:id_vis} shows the visualization of semantic encoding. We convert the image into discrete codes using the proposed SDE tokenizer, group the patches of the image according to their codes, and mark them with rectangular boxes. The left image indicates that the two IDs represent the cat's ears and the area near its nose, respectively. The right image visualizes the code that represents strawberries. 
The illustration demonstrates that the discrete codes extracted by the SDE tokenizer contain high-level semantic information, thus significantly enhancing the understanding capability (as shown in Table 6).

 \paragraph{Image Reconstruction}

\begin{figure}[!htbp]
  \centering
   \includegraphics[width=1\linewidth]{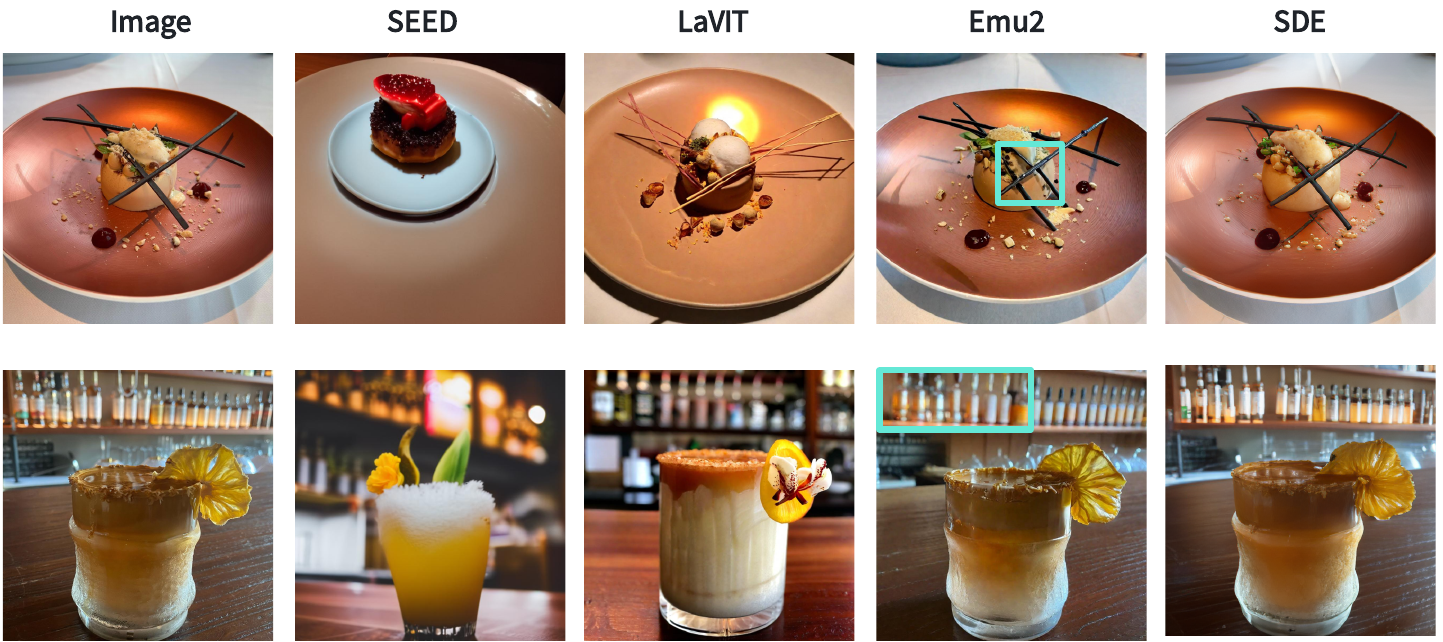}
   \caption{Comparison of image reconstruction results with different methods \cite{ge2023making,jin2024unified,sun2024generative}. The original image is in the first column, and SDE is the proposed tokenizer. }
   \label{fig:reconstruct_demo}
\end{figure}

 Figure \ref{fig:reconstruct_demo} shows the comparison of the image reconstruction results with other semantic tokenizers, where the first column is the original image. We observe that methods like SEED \cite{ge2023making} and LaVIT \cite{jin2024unified} can only retain basic semantic information, but show significant differences in color, number of objects, and background compared to the original image. Emu2 \cite{sun2024generative} failed to accurately restore some details (the rectangular box in the figure). 
The proposed tokenizer explicitly integrates high-level semantic information and low-level information during the discretization process, so the reconstructed results perform better in preserving both the major objects and the details.

\begin{figure}[htbp]
  \centering
   \includegraphics[width=1\linewidth]{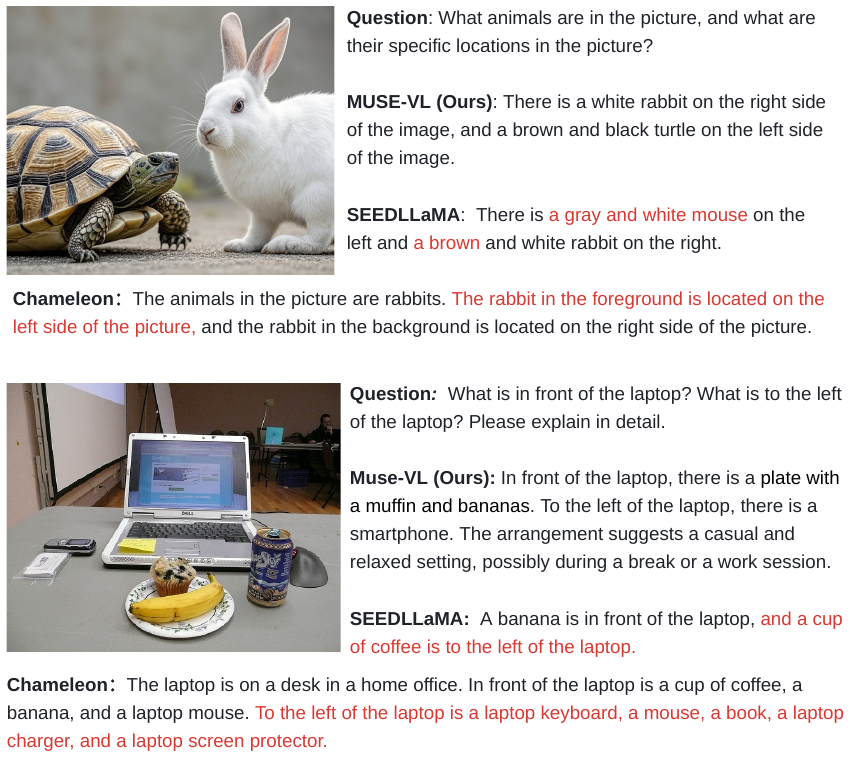}

   \caption{Comparison of results on the Visual Question Answering (VQA) task. The model is required to answer the user's questions based on the input image. The inaccurate parts of the response are highlighted in red. }
   \label{fig:demo}
\end{figure}

\paragraph{Visualization of VQA}
Figure \ref{fig:demo} illustrates MUSE-VL's ability to tackle visual question-answering tasks. The model receives an image as its initial input, after which the user poses questions regarding the image. 
The results show that the Chameleon \cite{team2024chameleon} and SEEDLLaMA \cite{ge2023making} models make obvious errors in animal recognition and spatial localization. Additionally, Chameleon describes objects that were not present in the image, indicating hallucination issues. Compared with them, the results show the proposed model can accurately answer questions based on image information, demonstrating that the model has effective spatial localization and instruction-following capabilities.

\begin{table}[!t]
\captionsetup{skip=2pt}
	\centering
    \caption{Evaluation on TextVQA benchmark.}
    \label{table_ocr}
    \begin{tabular}{llc}
    \toprule
Method      & Resolution         & TextVQA \\
\midrule
LLAVA 1.5 \cite{liu2024improved}   & 336                & 58.2    \\
Janus  \cite{wu2024janus}     & 384                & 50.7    \\
VILA-U \cite{wu2024vilau}     & 256                & 48.3    \\
VILA-U  \cite{wu2024vilau}    & 384                & 60.8    \\
EMU3  \cite{wang2024emu3}      & 1024               & 64.7    \\
SynerGen-VL \cite{li2025synergen} & Dynamic & 67.5    \\
\midrule
MUSE-VL      & 256                & 52.8    \\
MUSE-VL      & 384                & 61.3  \\

    \bottomrule
    \end{tabular}
\end{table}

\paragraph{Benchmarks of high-resolution benchmarks}
Table \ref{table_ocr} presents the results of the commonly used TextVQA benchmark. This benchmark is highly relevant to OCR tasks and therefore requires high-resolution image understanding capabilities. It is worth noting that our current model does not yet include the training process of high-resolution images. We plan to support high-resolution input in future work.

\section{Visual Generation Results}

Table \ref{fig:geneval} shows the quantitative results of the text-to-image in GenEval \cite{2023geneval} benchmark and compares them with other state-of-the-art generation models. We followed DALL-E 3 \cite{2023dalle3} to rewrite the prompts, making them more aligned with the dense captions in the training data. 

The results show that our model exhibits better performance than other unified models such as Chameleon \cite{team2024chameleon} and SEED-X \cite{ge2024seedx}. And it achieves performance close to the diffusion models. This indicates that our model has a strong image-text alignment capability.

\section{Limitation and Future Work}

Due to the limitations in the scale of training data and the resolution of generated images, our model has not surpassed the SOTA diffusion models in visual generation. In the future, we plan to further enhance the generation quality by expanding the scale of the training dataset for visual generation and using a more powerful image encoder \cite{tian2025var}. 
Furthermore, exploring the native integration of AR and Diffusion to further enhance the quality of image generation and instruction following is both challenging and promising.

In this work, extensive experiments and evaluations have been conducted on multimodal understanding and text-to-image tasks, demonstrating that our model can effectively unify the modeling of textual and visual data. Moreover, the architecture of our model supports arbitrary sequences of images and text. The next step is to further expand the capabilities of MUSE-VL by incorporating interleaved image-text data and image-editing data during training.

\end{document}